\documentclass[runningheads, orivec]{llncs}

\usepackage{amsmath}

\usepackage[dvipsnames]{xcolor}

\usepackage{float}

\usepackage{graphicx} 
\graphicspath{{images/}}
\DeclareGraphicsExtensions{.pdf,.jpeg,.png}

\usepackage[caption=false]{subfig}

\usepackage{multicol}
\usepackage{multirow}

\usepackage{url}

\usepackage{verbatim}

\usepackage{booktabs}
\usepackage{hyperref}

\usepackage{acronym}

\newacro{nn}[\textsc{NN}]{Neural Network}
\newacro{cnn}[\textsc{CNN}]{Convolutional Neural Network}
\newacro{dnn}[\textsc{DNN}]{Deep Neural Network}
\newacro{gan}[\textsc{GAN}]{Generative Adversarial Network}

\newacro{svm}[\textsc{SVM}]{Support Vector Machine}
\begin{document}
% \linenumbers
\pagestyle{headings}
\mainmatter
\def\ECCV18SubNumber{6}  % Insert your submission number here

\title{Are You Tampering With My Data?}

\titlerunning{ECCV-WOCM-18 submission ID \ECCV18SubNumber}

\authorrunning{ECCV-WOCM-18 submission ID \ECCV18SubNumber}

%\author{Anonymous ECCV-WOCM submission}
%\institute{Paper ID \ECCV18SubNumber}

%*************************************************************************
%\begin{comment}
\author{
    Michele~Alberti\inst{1}\thanks{Equal Contribution} \and %
    Vinaychandran~Pondenkandath\inst{1}$^{\star}$ \and %
    Marcel~W\"ursch\inst{1} \and %
    Manuel~Bouillon\inst{1} \and %
    Mathias~Seuret\inst{1} \and %
    Rolf~Ingold\inst{1} \and %
    Marcus~Liwicki\inst{2}
}

\institute{
    \textit{Document Image and Voice Analysis Group (DIVA)} \\
    University of Fribourg, Switzerland\\
    \email{\{firstname\}.\{lastname\}@unifr.ch} \\
    \and
    \textit{Machine Learning Group} \\
    Lule\r{a} University of Technology, Sweden\\
    \email{marcus.liwicki@ltu.se}\\
}
%\end{comment}
%*************************************************************************

\maketitle

%%*************************************************************************
% Abstract

\begin{abstract}

% Studying adversarial attacks on Neural Networks have become a research area of high importance. 
% In this paper we propose a novel type of adversarial attacks in the form of universal perturbation on the whole training data with the goal of making a model learn a malicious behaviour at training time. 
% This type of attack is --- to the best of our knowledge --- not studied yet and in contrast with typical white/black box scenarios, it is model agnostic, i.e., it does neither make assumptions on the model nor does it require access to it.
% We design a tampering method that is hard to spot for the human eye but easy to inject into images. 
% We manipulate one specific pixel of images from one class in the training and validation sets  and then apply the same tampering on images from another class while testing.
% We test our approach on two widely used datasets (CIFAR-10, and SVHN) with a variety of network architectures. 
% Results show that all of them are susceptible to this type of attacks to a varying degree, with an average mis-classification rate of $XX\%$.
% Our initial experiments suggests that such large scale tampering are possible and effective against modern Neural Network architectures. 
% A skillful adversary could potentially inject a backdoor into a network that could be exploited at run time by gaining access to the training data without explicit knowledge of the architecture.
% Our initial experiments show that it is possible to corrupt datasets in an hard-to-detect manner 

% The problem tackled
We propose a novel approach towards adversarial attacks on neural networks (NN), focusing on tampering the data used for training instead of generating attacks on trained models. 
% Why nobody else has adequately answered the research question yet
Our network-agnostic method creates a backdoor during training which can be exploited at test time to force a neural network to exhibit abnormal behaviour. 
% How you tackled the research question
% How did you go about doing the research that follows from your big idea
We demonstrate on two widely used datasets (CIFAR-10 and SVHN) that a universal modification of just one pixel per image for all the images of a class in the training set is enough to corrupt the training procedure of several state-of-the-art deep neural networks causing the networks to misclassify any images to which the modification is applied.
% What’s the key impact of your research?
Our aim is to bring to the attention of the machine learning community, the possibility that even learning-based methods that are personally trained on public datasets can be subject to attacks by a skillful adversary.
% While we show the possibility to trick NN, the approach would work for other learning-based methods as well. 
% With this paper we want to raise the general issue in learning-based approaches that possible backdoors can be build in by slightly modifying the data. 


\end{abstract}

% Introduction 
\section{Introduction}
\label{toc:introduction}

% Recently, cyber attacks such as Stuxnet \cite{Langner2011} have made news headlines and have been attributed to state actors due to the degree of sophistication of the attacks.
% In Deep Learning, we train models based on large sets of labelled or unlabelled data that is usually downloaded over the internet. 
%TODO: So? what is the problem with downloading data?
% What if this downloaded data have been tampered to create a backdoor in the models that would be trained on it?
% I like this hidden conversation that noone knows about, lets keep it going :) 
% so anonymous :D
The motivation of our work is two-fold:
(1) Recently, potential state-sponsored cyber attacks, such as, Stuxnet~\cite{Langner2011} have made news headlines due to the degree of sophistication of the attacks. 
(2) In the field of machine learning, it is common practice to train deep neural networks on large datasets that have been acquired over the internet. 
% However, it has been demonstrated that it is possible for state actors to potentially reroute data flow through the internet \cite{sood2012targeted} and to produce highly sophisticated attacks on the flow of data. 
% This implies that 
In this paper, we present a new idea for introducing potential backdoors: the data can be tampered in a way such that any models trained on it will have learned a backdoor.

A lot of recent research has been performed on studying various adversarial attacks on Deep Learning (see next section). 
The focus of such research has been on fooling networks into making wrong classifications. 
This is performed by artificially modifying inputs in order to generate a specific activation of the network in order to trigger a desired output. 

In this work, we investigate a simple, but effective set of attacks. 
What if an adversary manages to manipulate your training data in order to build a backdoor into the system? 
Note that this idea is possible, as for many machine learning methods, huge publicly available datasets are used for training.
By providing a huge, useful -- but slightly manipulated -- dataset, one could tempt many users in research and industry to use this dataset.
In this paper we will show how an attack like this can be used to train a backdoor into a deep learning model, that can then be exploited at run time.

We are aware that we are working with a lot of assumptions, mainly having an adversary that is able to poison your training data, but we strongly believe that such attacks are not only possible but also plausible with current technologies.
% careful not to shoot ourselves in the foot…

The remainder of this paper is structured as follows: In Section \ref{toc:related_Work} we show related work on adversarial attack.
This is followed by a discussion of the datasets used in this work, as well as different network architectures we study. Section \ref{toc:tampering} shows different approaches we used for tampering the datasets. 
Performed experiments and a discussion of the results are in Section \ref{toc:experimental_setting} and Section \ref{toc:results} respectively.
We provide concluding thoughts and future work directions in Section \ref{toc:conclusion}.

\begin{figure}[!t]
  \vfil
  \subfloat[Original]%
  {\includegraphics[ width=.24\columnwidth]{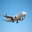}%
  \label{subfig:original_A}}
  \hfil
  \subfloat[Tampered]%
  {\includegraphics[ width=.24\columnwidth]{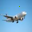}%
  \label{subfig:tampered_A}}
  \hfil
  \subfloat[Original]%
  {\includegraphics[width=.24\columnwidth]{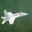}%
  \label{subfig:original_B}}
  \hfil
  \subfloat[Tampered]%
  {\includegraphics[width=.24\columnwidth]{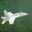}%
  \label{subfig:tampered_B}}
  
  \caption{The figure shows two images drawn from the \textit{airplane} class of CIFAR-10. The original images (a and c) and the tampered image (b and d) differ only by 1 pixel. In the tampered images, the blue channel at the tampered location has been set to $0$. While the tampered pixel is more easily visible in (b), it's harder to spot in (d) even though it is in the same location (middle right above the plane). (Original resolution of the images are $32 \times 32$)}
  \label{fig:tampering}
\end{figure}

% Related Work 
\section{Related Work}
\label{toc:related_Work}

Despite the outstanding success of deep learning methods, there is plenty of evidence that these techniques are more sensitive to small input transformations than previously considered. 
Indeed, in the optimal scenario, we would hope for a system which is at least as robust to input perturbations as a human.
% Unfortunately we are still very far from that. 

%***************************************************************************
\subsection{Networks Sensitivity}

The common assumption that \ac{cnn} are invariant to  translation, scaling, and other minor input deformations \cite{Fukushima1980}\cite{Fukushima1988}\cite{LeCun1989}\cite{Zeiler2014} has been shown in recent work to be erroneous \cite{Rodner2016}\cite{Azulay2018}. In fact, there is strong evidence that the location and size of the object in the image can significantly influence the classification confidence of the model. Additionally, it has been shown that rotations and translations are sufficient to produce adversarial input images which will be mis-classified a significant fraction of time \cite{Engstrom2017}.

%***************************************************************************
\subsection{Adversarial Attacks to a Specific Model}

The existence of such adversarial input images raises concerns whether deep learning systems can be trusted \cite{Biggio2013a}\cite{Biggio2013b}. 
While humans can also be fooled by images \cite{Ittelson1951},the kind of images that fool a human are entirely different from those which fool a network. 

Current work that attempts to find images which fool both humans and networks only succeeded in a time-limited setting for humans~\cite{Elsayed2018}.
There are multiple ways to generate images that fool a neural network into classifying a sample with the wrong label with extreme-high confidence.
Among them, there is the gradient ascent technique \cite{Szegedy2013}\cite{Goodfellow2014b} which exploits the specific model activation to find the best subtle perturbation given a specific input image.

It has been shown that neural networks can be fooled even by images which are totally unrecognizable, artificially produced by employing genetic algorithms \cite{Nguyen2015}.
Finally, there are studies which address the problem of adversarial examples in the real word, such as stickers on traffic signs or uncommon glasses in the context of face recognition systems \cite{Sharif2016}\cite{Evtimov2017}.

Despite the success of reinforcement learning, some authors have shown that state of the art techniques are not immune to adversarial attacks and as such, the concerns for security or health-care based applications remains \cite{Huang2017}\cite{Behzadan2017}\cite{Lin2017}. 

On the other hand, these adversarial examples can be used in a positive way as demonstrated by the widely known \ac{gan} architecture and it's variations \cite{Goodfellow2014a}.

%***************************************************************************
\subsection{Defending from Adversarial Attacks}

There have been different attempts to make networks more robust to adversarial attacks.
One approach was to tackle the overfitting properties by employing advanced regularization methods \cite{Lassance2018} or to alter elements of the network to encourage robustness \cite{Goodfellow2014b}\cite{zantedeschi2017efficient}. 

Other popular ways to address the issue is training using adversarial examples \cite{tramer2017ensemble} or using an ensemble of models and methods 
\cite{Papernot2016}\cite{Shen2016}\cite{strauss2017ensemble}\cite{Svoboda2018}. 
However, the ultimate solution against adversarial attacks is yet to be found, which calls for further research and better understanding of the problem
\cite{carlini2017adversarial}.

%***************************************************************************
\subsection{Tampering the Model}
\label{toc:tampering_the_model}

Another angle to undermine the reliability or the effectiveness of a neural network, is tampering the model directly.
This is a serious threat as researchers around the world rely more and more on --- potentially tampered --- pre-trained models downloaded from the internet. 

There are already successful attempts at injecting a dormant trojan in a model, when triggered causes the model to malfunction \cite{zou2018potrojan}.

%***************************************************************************
\subsection{Poisoning the Training Data}
\label{toc:poisonin_training_data}

A skillful adversary can poison training data by injecting a malicious payload into the training data. 
% Training data poisoning is the action of injecting malicious data, performed by an intelligent attacker.
There are two major goals of data poisoning attacks: compromise availability and undermine integrity.

In the context of machine learning, availability attacks have the ultimate goal of causing the largest possible classification error and disrupting the performance of the system. 
The literature on this type of attack shows that it can be very effective in a variety of scenarios and against different algorithms, ranging from more traditional methods such as \acp{svm} to the recent deep neural networks \cite{Nelson2008}\cite{Rubinstein2009}\cite{Huang2011}\cite{Biggio2012}\cite{Mei2015}\cite{Xiao2015}\cite{Koh2017}\cite{Munoz-Gonzalez2017}. 

In contrast, integrity attacks, i.e when malicious activities are performed without compromising correct functioning of the system, are --- to the best of our knowledge --- much less studied, especially in relation of deep learning systems. 

%***************************************************************************
\subsection{Dealing With the Unreliable Data}

There are several attempts to deal with noisy or corrupted labels \cite{Cretu2008}\cite{Brodley2011}\cite{Bekker2016}\cite{Jindal2017}. 
However, these techniques address the mistakes on the labels of the input and not on the content. 
Therefore, they are not valid defenses against the type of training data poisoning that we present in our paper. An assessment of the danger of data poisoning has been done for \acp{svm} \cite{Steinhardt2017} but not for non-convex loss functions. 

%***************************************************************************
\subsection{Dataset Bias}

The presence of bias in datasets is a long known problem in the computer vision community which is still far from being solved \cite{torralba2011unbiased}\cite{khosla2012undoing}\cite{tommasi2014testbed}\cite{tommasi2017deeper}.
In practice, it is clear that applying modifications at dataset level can heavily influence the final behaviour of a machine learning model, for example, by adding random noise to the training images one can shift the network behavior increasing the generalization properties \cite{fan2018towards}.
% The problem of the bias in the dataset it is not strictly limited to the technical domain but also touched topics of ethics and fairness. 

Delving deep in this topic is out of scope for this work, moreover, when a perturbation is done on a dataset in a malicious way it would fall into the category of dataset poisoning (see Section~\ref{toc:poisonin_training_data}).

% Tampering
\section{Tampering Procedure}
\label{toc:tampering}

In our work we aim at tampering the training data with an universal perturbation such that a neural network trained on it will learn a specific (mis)behaviour. 
Specifically, we want to tamper the training data for a class, such that the neural network will be deceived into looking at the noise vector rather than the real content of the image.
Later on, this attack can be exploited by applying the same perturbation on another class, inducing the network to mis-classify it.

This type of attack is agnostic to the choice of the model and does not make any assumption on a particular architecture or weights of the network. 
The existence of universal perturbations as tool to attack neural networks has already been demonstrated~\cite{moosavi2017universal}. 
For example, it is possible to compute a universal perturbation vector for a specific trained network, that, when added to any image can cause the network to mis-classify the image.
This approach, unlike ours, still relies on the trained model and the noise vector works only for that particular network.
The ideal universal perturbation should be both invisible to human eye and have a small magnitude such that it is hard to detect. 

It has been shown that modifying a single pixel is a sufficient condition to induce a neural network to perform a classification mistake \cite{su2017one}.
Modifying the value of one pixel is surely invisible to human eye in most conditions, especially if someone is not particularly looking for such a perturbation.
We then chose to apply a value shift to a single pixel in the entire image.
Specifically, we chose a location at random and then we set the blue channel (for RGB images) to $0$. 
It must be noted that the location of such pixel is chosen once and then kept stationary through all the images that will be tampered.

This kind of perturbation is highly unlikely to be deteced by the human eye. Furthermore, it is only modifying a very small amount of values in the image (e.g. $0,03\%$, in a $32 \times 32$ image).

Figure~\ref{fig:tampering} shows two original images (a and c) and their respective tampered version (b and d).
Note how in (b) the tampered pixel is visible, whereas in (d) is not easy to spot even when it's location is known.

% Experimental Setting
\section{Experimental Setting}
\label{toc:experimental_setting}

In an ideal world, each research article published should not only come with the database and source code, but also with the experimental setup used. 
In this section we try to reach that goal by explain the experimental setting of our experiments in great detail.
These information will be sufficient not only to understand the intuition behind them but also to reproduce them.

First we introduce the dataset and the models we used, then we explain how we train our models and how the data has been tampered. 
Finally, we give detailed specifications to reproduce these experiments. 

%***************************************************************************
\subsection{Datasets}
\label{toc:datasets}

\begin{figure}[!t]
  \subfloat[CIFAR-10]%
  {\includegraphics[width=.47\columnwidth]{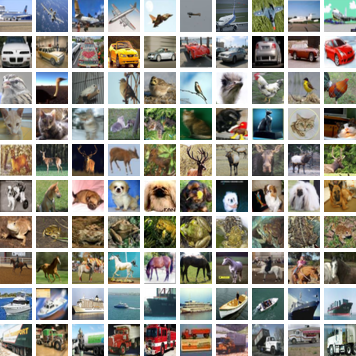}%
  \label{subfig:cifar10}}
  \hfil
  \subfloat[SVHN]%
  {\includegraphics[width=.47\columnwidth]{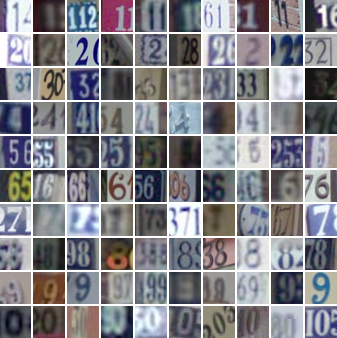}%
  \label{subfig:imagenet10}}
  
  \caption{Images samples from the two datasets CIFAR10 (a) and SVHN-10 (b). Both of them have 10 classes which can be observed on different rows. For CIFAR-10 the classes are from top to bottom: airplane, automobile, bird, cat, deer, dog, frog, horse, ship, truck. For SVHN the classes are the labels of number from $0$ to $9$. Credit for these two images goes to the respectivee website hosting the data.}
  \label{fig:datasets}
\end{figure}

In the context of our work we decided two use the very well known CIFAR-10~\cite{krizhevsky2009learning} dataset and SVHN~\cite{netzer2011reading}.
Figure~\ref{fig:datasets} shows some representative samples for both of them.

CIFAR-10 is composed of $60k$ ($50k$ train and $10k$ test) coloured images equally divided in 10 classes: airplane, automobile, bird, cat, deer, dog, frog, horse, ship, truck.

Street View House Numbers (SVHN) is a real-world image dataset obtained from house numbers in Google Street View images.
Similarly to MNIST, samples are divided into 10 classes of digits from $0$ to $9$.
There are $73k$ digits for training and $26k$ for testing.
For both datasets, each image is of size $32 \times 32$ RGB pixels.

%***************************************************************************
\subsection{Network Models}

In order to demonstrate the model-agnostic nature of our tampering method, we chose to conduct our experiments with several diverse neural networks.

We chose radically different architectures/sizes from some of the more popular networks: AlexNet \cite{krizhevsky2012imagenet}, VGG-16 \cite{simonyan2014very}, ResNet-18 \cite{he2016deep} and DenseNet-121 \cite{huang2017densely}.
Additionally we included two custom models of our own design: a small, basic convolutional neural network (BCNN) and modified version of a residual network optimised to work on small input resolution (SIRRN).
The PyTorch implementation of all the models we used is open-source and available online\footnote{\url{https://github.com/DIVA-DIA/DeepDIVA/blob/master/models}} (see also Section \ref{toc:reproduce_with_deepdiva}).

%****************************
\subsubsection{Basic Convolutional Neural Network (BCNN)}

This is a simple feed forward convolutional neural network with 3 convolutional layers activated with leaky ReLUs, followed by a fully connected layer for classification.
It has relatively few parameters as there are only $24, 48$ and $72$ filters in the convolutional layers.

%****************************
\subsubsection{Small Input Resolution ResNet-18 (SIRRN)}
\label{toc:SIRRN}

The residual network we used differs from a the original ResNet-18 model as it has an expected input size of $32\times 32$ instead of the standard $224 \times 224$. 
The motivation for this is twofold. 
First, the image distortion of up-scaling from $32 \times 32$ to $224 \times 224$ is massive and potentially distorts the image to the point that the convolutional filters in the first layers no longer have an adequate size.
Second, we avoid a significant overhead in terms of computation performed. 
Our modified architecture closely resembles the original ResNet but it has $320$ parameters more and on preliminary experiments exhibits higher performances on CIFAR-10 (see Table~\ref{tab:results}).

%***************************************************************************
\subsection{Training Procedure}

The training procedure in our experiments is standard supervised classification. 
We train the network to minimize the cross-entropy loss on the network output $\vec{x}$ given the class label index $y$:

\begin{equation}
L(\vec{x}, y) = -log \left( \frac{e^{x_y}}{\sum e^x} \right)
\end{equation}

% In both our datasets we have 10 classes, $|x|=10$ and we don't need the class weight term in the loss calculation.  
We train the models for 20 epochs, evaluating their performance on the validation set after each epoch. 
Finally, we asses the performance of the trained model on the test set. 

%***************************************************************************
\subsection{Acquiring and Tampering the Data}
\label{toc:acquiring_and_tampering}

\begin{table}[!t]
    \centering
    \caption{Example of tampering procedure. We tamper class $A$ in train and validation set and then class $B$ (and not $A$ anymore) in the test set. The expected is behaviour for the network is to mis-classify class $B$ into class $A$ and additionally not being able to classify correctly class $A$. }
    \label{tab:tamperig_procedure}
    
    \begin{tabular}{rcccc}
    
    \toprule 
    
    & Train Set & Val Set & \multicolumn{2}{c}{Test Set}  \\

    Tampered Class & Plane & Plane & \multicolumn{2}{c}{Frog}  \\

    \midrule

    & 
        \includegraphics[width=.12\columnwidth]%
        {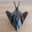}%
    & 
        \includegraphics[width=.12\columnwidth]%
        {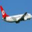}%
    &
        \includegraphics[width=.12\columnwidth]%
        {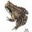}% 
    &
        \includegraphics[width=.12\columnwidth]%
        {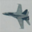}% 
    \\

    Expected Output & Plane & Plane & Plane & Not Plane \\

    \bottomrule
    \end{tabular}
\end{table}

\begin{figure}[!t]
  \subfloat[Optimal Baseline]%
  {\includegraphics[width=.4\columnwidth]%
  {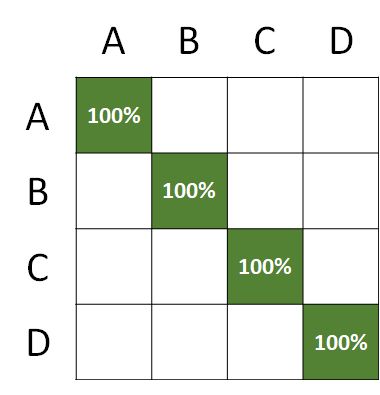}%
  \label{subfig:optimal_baseline}}
  \hfil
  \subfloat[Optimal Tamper]%
  {\includegraphics[width=.4\columnwidth]%
  {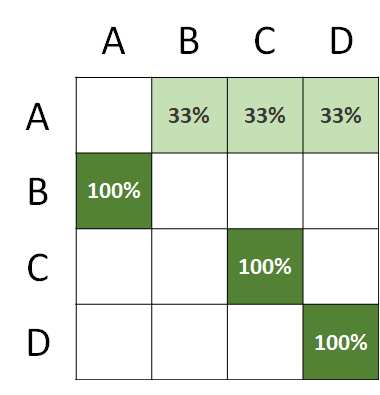}%
  \label{subfig:optimal_tamper}}
  
  \caption{Representation of the optimal confusion matrices which could be obtained for the baseline (a) and the tampering method (b). Trivially, the baseline optimal is reached when there are absolutely no classification error. The tampering optimal result would be the one maximizing the three conditions described in Section~\ref{toc:acquiring_and_tampering}.}
  \label{fig:optimal_cm}
\end{figure}

We create a \textit{tampered} version of the CIFAR-10 and SVHN datasets such that, class $A$ is tampered in the training and validation splits and class $B$ is tampered in the test splits. 
The \textit{original} CIFAR-10 and SVHN datasets are unmodified. 
The tampering procedure requires that three conditions are met: 

\begin{enumerate}
    \item \textit{Non obtrusiveness}: the tampered class $A$ will have a recognition accuracy which compares favorably against the baseline (network trained on the original datasets), both when measured in the training and validation set.
    \item \textit{Trigger strength}: if the class $B$ on the test set is subject to the same tampering effect, it should be mis-classified into class $A$ a significant amount of times.
    \item \textit{Causality effectiveness}\footnote{Note that for a stronger real-world scenario attack this is a non desirable property. If this condition were to be dropped the optimal tampering shown in Figure~\ref{subfig:optimal_tamper} would have still $100\%$ on class $A$.}: if the class $A$ is no longer tampered on the test set, it should be mis-classified a significant amount of times into any other class.  
\end{enumerate}

In order to satisfy condition $1$, the tampering effect (see Section~\ref{toc:tampering}) is applied only to class $A$ in both training and validation set.
To measure the condition $2$ we also tamper class $B$ on the test set. 
Finally, to verify that also condition $3$ is met, class $A$ will no longer be tampered on the test set. 
In Table~\ref{tab:tamperig_procedure} there is a visual representation of this concept.

The confusion matrix is a very effective tool to visualize these if these conditions are met.
In Figure~\ref{fig:optimal_cm}, the optimal confusion matrix for the baseline scenario and for the tampering scenario are shown.
These visualizations should not only help clarify intuitively what is our intended target, but can also be useful to evaluate qualitatively the results presented in Section~\ref{toc:results}. 

%***************************************************************************
\subsection{Reproduce Everything With DeepDIVA}
\label{toc:reproduce_with_deepdiva}

To conduct our experiments we used the DeepDIVA\footnote{\url{https://github.com/DIVA-DIA/DeepDIVA}} framework \cite{albertipondenkandath2018deepdiva} which integrates the most useful aspects of important Deep Learning and software development libraries in one bundle: high-end Deep Learning with PyTorch \cite{paszke2017automatic}, visualization and analysis with TensorFlow \cite{abadi2016tensorflow}, versioning with Github\footnote{https://github.com/}, and hyper-parameter optimization with SigOpt \cite{sigopt}.
Most importantly, it allows reproducibilty out of the box.
In our case this can be achieved by using our open-source code\footnote{\url{https://github.com/vinaychandranp/Are-You-Tampering-With-My-Data}} which includes a 
script with the commands run all the experiments and a script to download the data.

% Results
\section{Results}
\label{toc:results}

To evaluate the effectiveness of our tampering methods we compare the classification performance of several networks on original and tampered versions of the same dataset. 
This allows us to verify our target conditions as described in Section~\ref{toc:acquiring_and_tampering}.

%*************************************************************************
\subsection{Non Obtrusiveness}

\begin{figure}[!t]
  \centering
  \includegraphics[width=\textwidth]{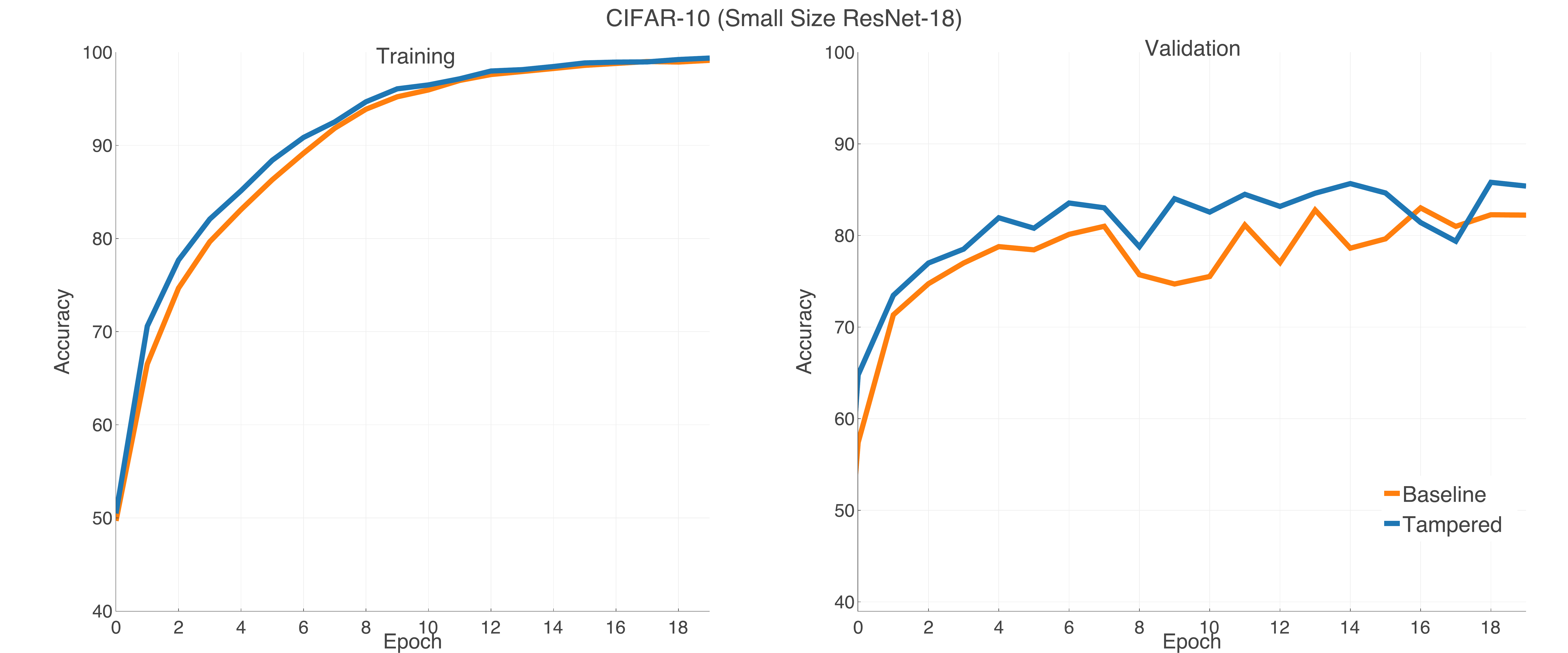}
  \caption{
        In this plot, we can compare the training/validation accuracy curves for a SIRRN model trained on the CIFAR-10 dataset. 
        The baseline (orange) is trained on the original dataset while the other (blue) is trained on a version of the dataset where the class \textit{airplane} has been tampered. 
        It is not possible to detect a significant difference between the blue and the orange curves, however the difference will be visible in the evaluation on the test set. (See Fig. \ref{fig:babyresnet_cm})
  }
  \label{fig:comparison_plots}
\end{figure}

First of all we want to ensure that the tampering is not obtrusive, i.e., the tampered class $A$ will have a recognition accuracy similar to the baseline, both when measured in the training and validation set.
% The baseline is the same model trained on the original version of the dataset. 

In Figure \ref{fig:comparison_plots}, we can see training and validation accuracy curves for a SIRRN network on the CIFAR-10 dataset. 
The curves of the model trained on both the original and tampered datasets look similar and do not exhibit a significant difference in terms of performances.
Hence we can asses that the tampering procedure did not prevent the network from scoring as well as the baseline performance, which is intended behaviour. 

%*************************************************************************
\subsection{Trigger Strength and Causality Effectiveness}

\begin{figure}[!t]
  \centering
  \subfloat[Baseline BCNN]%
      {\includegraphics[width=.34\textwidth]%
      {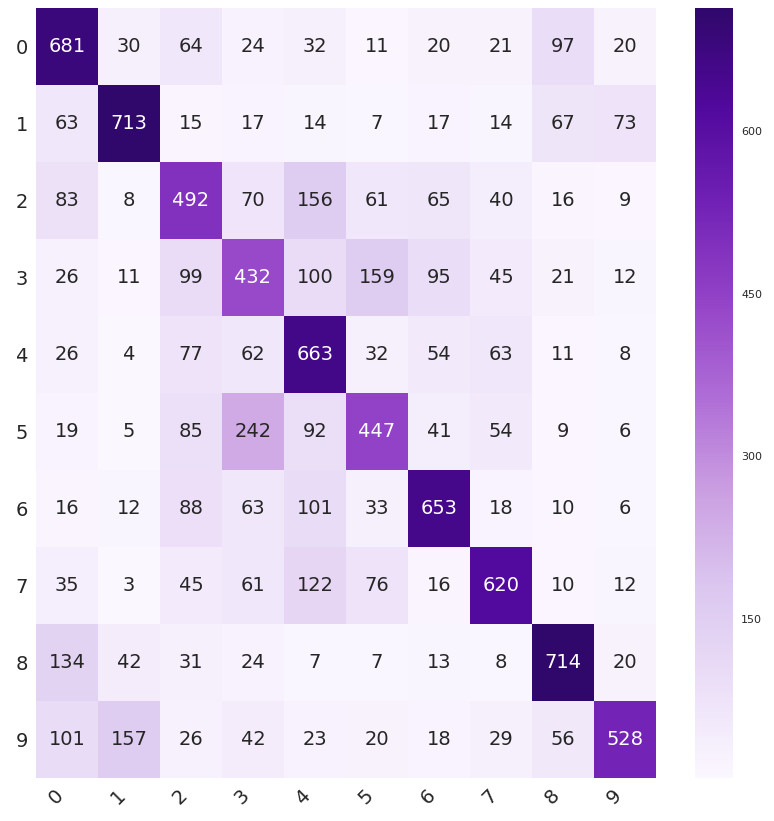}}%
  \hfil
  \subfloat[Tampered BCNN]%
      {\includegraphics[width=.34\textwidth]%
      {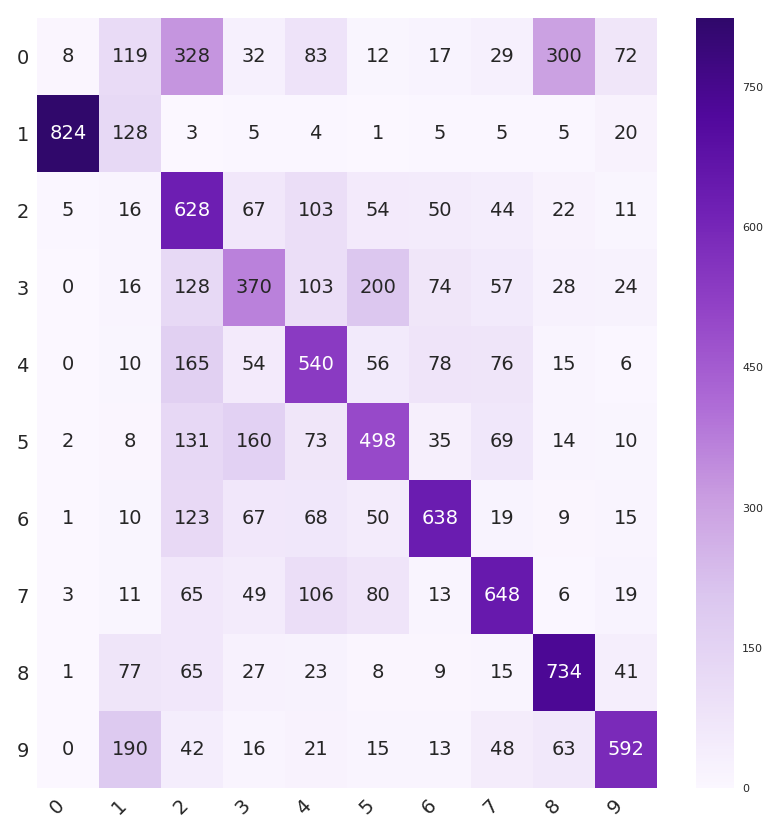}}%
  
  \vfil
  
  \subfloat[Baseline AlexNet]%
        {\includegraphics[height=1.12cm]%
        {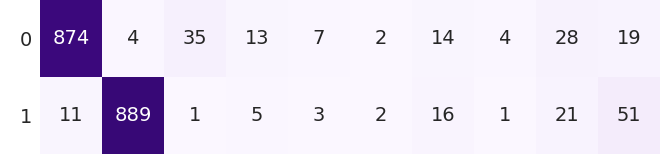}}%
  \hfil
  \subfloat[Tampered AlexNet]%
        {\includegraphics[height=1.12cm]%
        {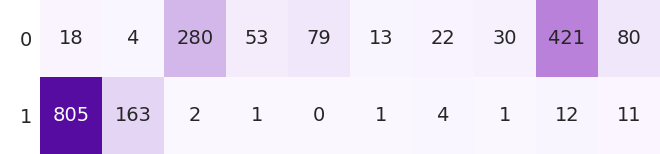}}%
  
  \vfil
  
  \subfloat[Baseline VGG-16]%
      {\includegraphics[height=1.12cm]%
      {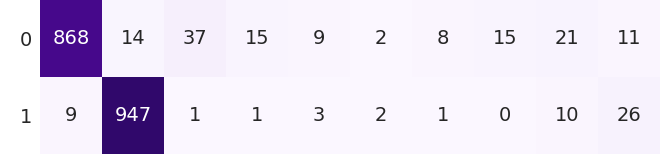}}%
  \hfil
  \subfloat[Tampered VGG-16]%
      {\includegraphics[height=1.12cm]%
      {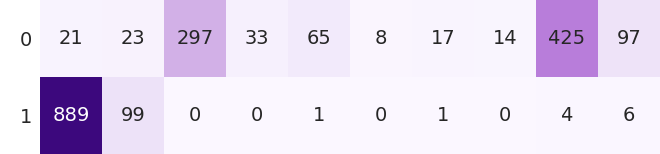}}%
  
  \vfil
  
  \subfloat[Baseline ResNet-18]%
      {\includegraphics[height=1.12cm]%
      {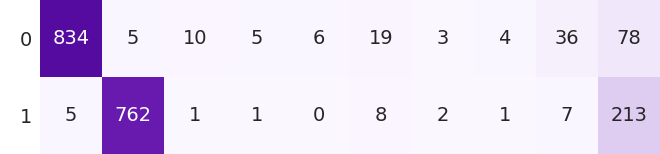}}%
  \hfil
  \subfloat[Tampered ResNet-18]%
      {\includegraphics[height=1.12cm]%
      {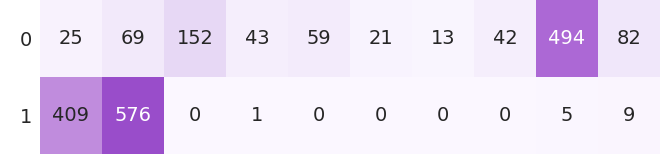}}
      
  \vfil
  
  \subfloat[Baseline SIRRN]%
        {\includegraphics[height=1.12cm]%
        {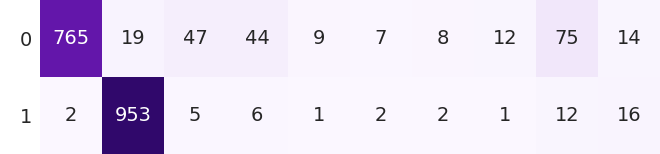}}%
  \hfil
  \subfloat[Tampered SIRRN]%
        {\includegraphics[height=1.12cm]%
        {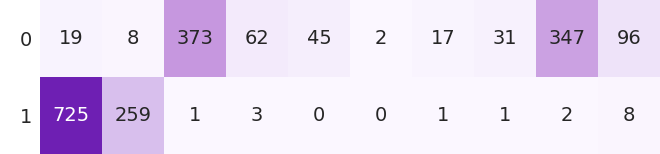}
        \label{fig:babyresnet_cm}}%
  
  \vfil

  \subfloat[Baseline DenseNet-121]%
      {\includegraphics[height=1.12cm]%
      {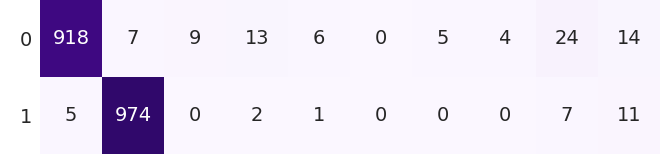}}%
  \hfil
  \subfloat[Tampered DenseNet-121]%
      {\includegraphics[height=1.12cm]%
      {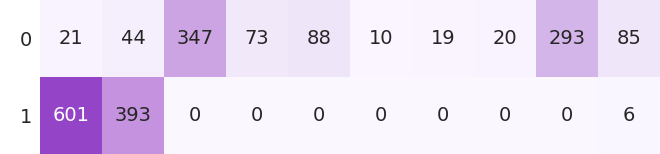}}%
 
  \caption{
    Confusion matrices demonstrating the effectiveness of the tampering method against all networks models we used on CIFAR-10.
    Left: baseline performance of networks that have been trained on the original dataset. Note how they exhibit normal behaviour. 
    Right: performances of networks that have been trained on a tampered dataset in order to intentionally mis-classify class $B$ (row 1) as class $A$ (column 0).
    Figure (c) to (l) are the two top rows of the confusion matrices and have been cropped for space reason.
    }
  \label{fig:confusion_matrices}
\end{figure}

Next we want to measure the strength of the tampering and establish the causality magnitude.
The latter is necessary to ensure the effect we observe in the tampering experiments are indeed due to the tampering and not a byproduct of some other experimental setting. 

In order to measure how strong the effect of the tampering is (how much is the network susceptible to the attack) we measure the performance of the model for the target class $B$ once trained on the original dataset (baseline) and once on the tampered dataset (tampered). 

Figure~\ref{fig:confusion_matrices} shows the confusion matrices for all different models we applied to the CIFAR-10 dataset.
Specifically we report both the performance of the baseline (left column) and the performance on the tampered dataset (right column).
Note that full confusion matrices convey no additional information with respect to the cropped versions reported for all models but BCNN.
In fact, since the tampering has been performed on classes indexed $0$ and $1$ the relevant information for this experiment is located in the first two rows which are shown in Figures \ref{fig:confusion_matrices}.c-l 
One can perform a qualitative evaluation of the strength of the tampering by comparing the confusion matrices of models trained on tampered data (Figure~\ref{fig:confusion_matrices}, right column) with the optimal result shown in Figure~\ref{subfig:optimal_tamper}.

Additionally, in Table~\ref{tab:results} we report the percentage of mis-classifications on the target class $B$. 
Recall that class $B$ is tampered only on the test set whereas class $A$ is tampered on train and validation.

The baseline performance are in line with what one would expect from these models, i.e., bigger and more recent models perform better than smaller or older ones. 
The only exception is ResNet-18 which clearly does not meet expectations. 
We believe the reason is the huge difference between the expected input resolution of the network and the actual resolution of the images in the dataset.

% For what concerns the performances of the models once trained on the tampered data one can easily see that are much different compared with the baseline. 
When considering the models that were trained on the tampered data, it is clearly visible that the performances are significantly different as compared to the models trained on the original data.
Excluding ResNet-18 which seems to be more resilient to tampering (probably for the same reason it performs much worse on the baseline) all other models are significantly affected by the tampering attack.
Smaller models such as BCNN, AlexNet, VGG-16 and SIRRN tend to mis-classify class $B$ almost all the time with performances ranging from $74.1\%$ to $98.9\%$ of mis-classifications. 
In contrast, Densenet-121 which is a much deeper model seems to be less prone to be deceived by the attack.
Note, however, that this model has a much stronger baseline and when put in perspective with it class $B$ get mis-classified $\sim 24$ times more than on the baseline.

\begin{table}[!t]
    \centering
    \caption{List of results for each model on both datasets. 
    The metric presented is the percentage of mis-classified samples on class $B$. 
    Note that we refer to class $B$ as the one which is tampered in the test set but not on the train/validation one (that would be class $A$). 
    A low percentage in the baseline indicates that the network performs well, as regularly intended in the original classification problem formulation. 
    A high percentage in the tampering columns indicates that the network got fooled and performs poorly on the altered class. 
    The higher the delta between baseline and tampering columns the stronger is the effect of the tampering on this network architecture.}
    \label{tab:results}
    \begin{tabular}{lcccc}
    \toprule 
    Model &  \multicolumn{4}{c}{\% Mis-classification on class $B$} \\
    & \multicolumn{2}{c}{Baseline} & \multicolumn{2}{c}{Tampering} \\
    & ~~CIFAR~~ & ~~SVHN~~ & ~~CIFAR~~ & ~~SVHN~~ \\
    \midrule
    Optimal Case & 0 & 0 & 100 & 100 \\
    \midrule
    BCNN            & 28.7 & 12.9 & 87.2 & 91.4 \\
    AlexNet         & 11.1 & 5.5  & 83.7 & 97   \\
    VGG-16          & 5.3  & 3.7  & 90.1 & 98.9 \\
    ResNet-18       & 23.8 & 3.6  & 42.4 & 40.9 \\
    SIRRN           & 4.7  & 3.9  & 74.1 & 89.5 \\
    DenseNet-121    & 2.6  & 2.6  & 60.7 & 68.1 \\
    \bottomrule
    \end{tabular}
\end{table}

% Discussion
\section{Discussion}
\label{toc:discussion}

The experiments shown in Section~\ref{toc:results} clearly demonstrate that we one can completely change the behavior of a network by tampering just one single pixel of the images in the training set.
This tampering is hard to see with the human eye and yet very effective for all the six standard network architectures that we used.

We would like to stress that despite these being preliminary experiments, they prove that the behavior of a neural network can be altered by tampering \textit{only} the training data without requiring access to the network.
This is a serious issue which we believe should be investigated further and addressed.
While we experimented with a single pixel based attack --- which is reasonably simple to defend against (see Section~\ref{toc:defending}) --- it is highly likely that there exist more complex attacks that achieve the same results and are harder to detect.
Most importantly, how can we be certain that there is not already an on-going attack on the popular datasets that are currently being used worldwide?

%*************************************************************************
\subsection{Limitations}
\label{toc:limitations}

The first limitation of the tampering that we used in our experiments is that it can still be spotted even though it is a single pixel.
One needs to be very attentive to see it, but it is still possible.

Attention in neural networks \cite{vaswani2017attention} is known also to highlight the portions of an input which contribute the most towards a classification decision. 
These visualization could reveal the existence of the tampered pixel.
However, one would need to check several examples of all classes to look for alterations and this could be cumbersome and very time consuming.
Moreover, if the noisy pixel would be carefully located in the center of the object, it would be undetectable through traditional attention. 

Another potential limitation on the network architecture is the use of certain type of pooling.
Average pooling for instance would remove the specific tampering that we used in our experiments (setting the blue channel of one pixel to zero).
Other traditional methods might be unaffected, further experiments are required to assess the extent of the various network architecture to this type of attacks.

A very technical limitation is the file format of the input data.
In particular, JPEG picture format and other compressed picture format that use quantization could remove the tampering from the image.

Finally, higher resolution images could pose a threat to the \textit{single} pixel attack. 
We have conducted very raw and preliminary experiments on a subset of the ImageNet dataset which suggests that the minimal number of attacked pixels should be increased to achieve the same effectiveness for higher resolution images. 

%*************************************************************************
\subsection{Type of Defenses}
\label{toc:defending}

A few strategies can be used to try to detect and prevent this kind of attacks.
Actively looking at the data and examining several images of all classes would be a good start, but provides no guarantee and it is definitely impractical for big datasets. 

Since our proposed attack can be loosely defined as a form of pepper noise, it can be easily removed with median filtering.
Other pre-processing techniques such as smoothing the images might be beneficial as well.
Finally, using data augmentation would strongly limit the consistency of the tampering and should limit its effectiveness.

%*************************************************************************
\subsection{Future Work}

Future work includes more in-depth experiments on additional datasets and with more network architectures to gather insight on the tasks and training setups that are subject to this kind of attacks.

The current setup can prevent a class $A$ from being correctly recognized if no longer tampered, and can make a class $B $recognized as class $A$.
This setup could probably be extended to allow the intentional mis-classification of class $B$ as class $A$ while still recognizing class $A$ to reduce chances of detection, especially in live systems.

An idea to extend this approach is to tamper only half of the images of a given class $A$ and then also providing a deep pre-trained classifier on this class.
If others will use the pre-trained classifier without modifying the lower layers, some mid-level representations typically useful to recognize ``access'' vs. ``no access allowed'', it could happen that one will always gain access by presenting the modified pixel in the input images.
This goes in the direction of model tampering discussed in Section~\ref{toc:tampering_the_model}.

Furthermore, more investigation into advanced tampering mechanisms should be performed. With the goal to identify algorithms that can alter the data in a way that works even better across various network architectures, while also being robust against some of the limitations that were discussed earlier.

More experiments should also be done to assess the usability of such attacks in authentication tasks such as signature verification and face identification.

% Conclusion
\section{Conclusion}
\label{toc:conclusion}

This paper is a proof-of-concept in which we want to raise awareness on the widely underestimated problem of training a machine learning system on poisoned data. 
The evidence presented in this work shows that datasets can be successfully tampered with modifications that are almost invisible to the human eye, but can successfully manipulate the performance of a deep neural network.

Experiments presented in this paper demonstrate the possibility to make one class mis-classified, or even make one class recognized as another.
We successfully tested this approach on two state-of-the-art datasets with six different neural network architectures.

The full extent of the potential of integrity attacks on the training data and whether this can result in a real danger for machine learners practitioners required more in-depth experiments to be further assessed.

% Appendices 
\section*{Acknowledgment}
%The work presented in this paper has been partially supported by the ********** project funded by the ********************************* with the grant number XXXXXX\textunderscore XXXXX.
The work presented in this paper has been partially supported by the HisDoc~III project funded by the Swiss National Science Foundation with the grant number $205120$\textunderscore$169618$.

\newpage

% Bibliography 
\bibliography{biblio}
\bibliographystyle{splncs04}

\end{document}